\documentclass[acmsmall,screen]{acmart}

\AtBeginDocument{%
  }

\begin{document}

\title{SetBERT: Enhancing Retrieval Performance for Boolean Logic and Set Operation Queries }

\author{Quan Mai, Susan Gauch, Douglas Adams}
\email{quanmai, sgauch, djadams@uark.edu}
\affiliation{%
  \institution{University of Arkansas}
  \city{Fayetteville}
  \state{Arkansas}
  \country{USA}
  \postcode{72701}
}


\renewcommand{\shortauthors}{Mai et al.}

\begin{abstract}
We introduce SetBERT, a fine-tuned BERT-based model designed to enhance query embeddings for set operations and Boolean logic queries, such as Intersection (AND), Difference (NOT), and Union (OR). SetBERT significantly improves retrieval performance for logic-structured queries, an area where both traditional and neural retrieval methods typically underperform. We propose an innovative use of inversed-contrastive loss, focusing on identifying the negative sentence, and fine-tuning BERT with a dataset generated via prompt GPT. Furthermore, we demonstrate that, unlike other BERT-based models, fine-tuning with triplet loss actually degrades performance for this specific task. Our experiments reveal that SetBERT-base not only significantly outperforms BERT-base (up to a 63\% improvement in Recall) but also achieves performance comparable to the much larger BERT-large model, despite being only one-third the size.
\end{abstract}

\ccsdesc[300]{Information systems~Novelty in information retrieval}
\ccsdesc[500]{Information systems~Language models}
\ccsdesc[300]{Information systems~Question answering}
\ccsdesc[300]{Information systems~Recommender systems}

\keywords{BERT, Neural Information Retrieval, Boolean Retrieval, Negation Retrieval.}


\maketitle

\section{Introduction}
Queries that involve set operations, such as Intersection (AND), Difference (NOT), and Union (OR) \cite{malaviya2023quest}, provide a convenient and efficient way to search for information, particularly when user requirements are highly specific or particularly complex. For instance, a movie enthusiast searching for films set in Vietnam but not centered on war might wish to query ``Vietnam movies not about war'' rather than a having to form a more specific search that contains a laundry list non-war-related  topics. Similarly, a student researching unique species indigenous to both Brazil and Mexico could save time by querying ``animals that only live in Brazil and Mexico'' instead of conducting separate searches and manually finding the intersection. These examples illustrate the power and utility of Boolean queries with implicit set operations to provide access to required information, streamlining the search process.

Both traditional and neural retrieval methods currently struggle with these types of queries \cite{malaviya2023quest}. Traditional retrieval methods, such as lexicon matching, often fail when faced with queries containing exclusions like ``not.'' For example, if a user is searching for documents on a particular topic but wants to exclude certain entities, traditional methods may not accurately filter out these undesired entities. This is because such methods typically rely on word presence detection without adequately understanding the context in which the words are used. Consequently, documents that contain the entities the user wants to exclude are often returned, leading to inaccurate search results.

Neural approaches, though more advanced and sophisticated, also have their share of limitations. For instance, BERT \cite{devlin2018bert} tends to treat many word pairs, including `and' and `not', as synonyms \cite{jeretic2020natural}. Consequently, BERT-based encoders perform poorly on queries containing these terms, leading to misinterpretation of the semantic differences between them. Similarly, T5-based encoders \cite{raffel2020exploring} also struggle with such queries \cite{malaviya2023quest}. This misinterpretation causes the embeddings of a query containing exclusions and a document containing the excluded entities to have a high similarity score, such as cosine similarity, because the encoder does not correctly understand the exclusion.

The current challenges faced by both traditional and neural retrieval methods in accurately interpreting and processing Boolean logic in queries highlight the need for more advanced retrieval techniques. In this paper, we introduce SetBERT, a fine-tuned BERT-based model trained on a synthesized dataset designed to enhance query embeddings for set operations and Boolean logic—specifically, Intersection (AND), Difference (NOT), and Union (OR). Our results demonstrate that the fine-tuned BERT significantly improves retrieval performance for these logic-structured queries. The code and datasets will be made available upon acceptance of this paper.

\section{Related Work}
\textbf{The Boolean retrieval model} The Boolean retrieval model has long been a cornerstone in the field of information retrieval (IR). This method utilizes Boolean logic to match documents to user queries by identifying whether specific terms are present or absent in a given text. Over the years, the model has undergone significant enhancements through contributions such as the Extended Boolean Model \cite{salton1983extended}, the InQuery system \cite{callan1995trec}, and the P-norm model \cite{fox1994combination}. These advancements have substantially increased the flexibility and effectiveness of Boolean retrieval methods, consolidating their place as widely used tools in the IR landscape.

However, despite the model's simplicity and ease of implementation, the traditional Boolean retrieval approach has notable shortcomings. A significant limitation is its inability to cater to partial matches, such as distinguishing between ``science-fiction'' and ``sci-fi''. The model also struggles with handling synonymy and polysemy—words that have multiple meanings or those that are synonymous. Currently, users must manually include all possible synonyms when formulating their queries, adding a layer of complexity and effort that could be mitigated with more sophisticated models.

\noindent\textbf{Neural Retrieval Methods,} such as those based on BERT \cite{devlin2018bert} and T5 \cite{raffel2020exploring}, promise to address some of the limitations of traditional Boolean models. These models leverage contextual embeddings to better understand the semantic nuances of queries and documents. Despite their advancements, neural models often falter when dealing with Boolean logic. For instance, BERT tends to treat words like `and' and `not' as near-synonyms \cite{jeretic2020natural}, leading to poor performance on queries involving these terms. Similarly, T5-based models exhibit comparable struggles \cite{malaviya2023quest}, primarily because they fail to properly interpret the semantic differences when Boolean logic is involved.

\noindent\textbf{The QUEST dataset} The introduction of the QUEST dataset for entity-seeking queries with implicit set operations \cite{malaviya2023quest} has greatly influenced our work. QUEST provides a critical benchmark for evaluating the capability of retrieval systems to handle queries involving implicit set operations. Constructed semi-automatically using Wikipedia category names and validated by crowdworkers for naturalness and relevance, QUEST offers a rigorous testbed. It challenges models to match multiple constraints and perform set operations accurately, highlighting the difficulties that modern systems face, especially with negation and conjunction operations.

Our work on SetBERT aims to bridge these gaps by enhancing the capabilities of BERT-based models to accurately interpret and process logic-structured queries. By fine-tuning a BERT model on a synthesized dataset designed to include set operations and Boolean logic, we provide a framework that significantly improves retrieval performance for these types of queries. Named in tribute to the work of QUEST, SetBERT represents a significant step towards more sophisticated and reliable retrieval methods. The effectiveness of SetBERT is evaluated against the rigorous benchmarks provided by datasets like QUEST, demonstrating its potential to advance the state-of-the-art in IR.

\section{Models struggle with Difference (NOT) and Intersection (AND)}
According to QUEST's evaluation results, intersections cause the most trouble and differences also pose difficulties for their strongest model. False positive errors are common in the case of intersections, where at least one constraint is satisfied. For instance, if the query is ``Flora that can only be found in South East Asia and China'', the erroneously retrieved documents often contain either ```Flora that can be found in South East Asia'' or ``Flora that can be found in China'', rather than flora that is exclusively found in both of these locations. This discrepancy highlights the challenges that intersections can pose for accurate data retrieval. Therefore, when generating a dataset for fine-tuning, we categorize sentences with only one constraint as negative samples. In contrast, sentences that include both constraints, or their paraphrases or synonyms, are classified as positive samples.

\textit{In case of negation, if users do not wish to retrieve an entity, place it at the bottom of the list.
} When dealing with queries that include exclusions, the range of possible answers expands considerably. This is because users are specifically seeking to omit, or exclude, certain entities from their search results. For instance, let us use the example of movie-goers wanting to watch films about Vietnam, but with an explicit exclusion of war-related themes. Here, there exist countless non-war themes that could be retrieved, including genres such as romance, comedy, drama, and more. In such contexts, the application of traditional contrastive loss becomes significantly challenging. Traditional contrastive loss seeks to maximize the similarity score between the query and positive document embeddings. However, when users express a desire to exclude certain entities, this traditional approach is not as effective.

We aim to avoid overfitting the model by not overloading it with numerous positive examples. Furthermore, covering all positive examples for queries involving the NOT operator is impractical. To address this challenge, our focus shifts to minimizing the similarity between the query and the negative documents. These negative documents explicitly contain entities that users have indicated they do not wish to retrieve. By reducing the similarity between the query and these negative documents, we can more effectively respect users' exclusion preferences, thus providing search results that better align with their specific requirements.

Additionally, this training approach benefits AND logic queries. By prioritizing documents that satisfy all constraints and relegating those with insufficient constraints to the bottom of the list, we mitigate the risk of retrieving documents that meet only one or two criteria. This strategy ensures that the retrieved documents are more comprehensive and relevant to the user's query.
\section{SetBERT}
\subsection{Data Generation}

\begin{figure}[h]
  \centering
  \includegraphics[width=\linewidth]{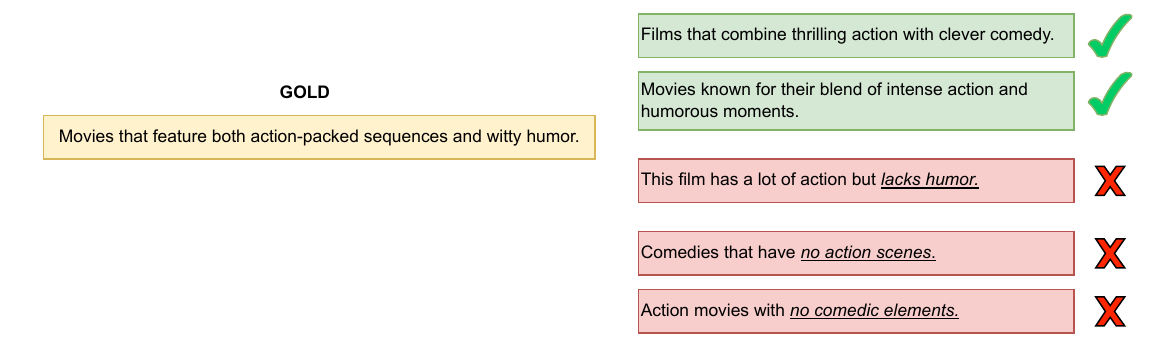}
  \caption{A generated sample for intersection (AND). The gold (or anchor) sentence is in yellow, positive sentences are in green, negative sentences are in red.}
  \Description{What a training sample looks like.}
  \label{fig: examples}
\end{figure}

We use OpenAI's \textit{gpt-3.5-turbo} to generate data based on specific prompts, each corresponding to a different type of Boolean logic. For a full list of these prompts, please refer to the appendix \ref{apd: PROMPT}. We have collected 50,000 samples for each operation, yielding a total of 150,000 samples. Each sample follows this format: an anchor (gold) sentence, a list of positive sentences, and a list of negative sentences. The gold sentence mimics a query containing only one set operation; we do not combine multiple operations in a single sentence. Positive sentences can either be directly interpreted from the gold sentence or generated by modifying the entities in the gold sentence with their synonyms or paraphrases. Negative sentences, however, lack the necessary entities found in the gold sentence. An example of a gold-positive-negative sample is shown in Figure \ref{fig: examples}.

\subsection{Fine-tuning BERT}
\subsubsection{Is triplet loss a good choice for fine-tuning SetBERT?} The objective of triplet learning is to ensure that the embeddings of positive sentences ($p^+$) closely resemble those of the anchor sentence ($p$), while maintaining a low similarity between the anchor and the negative sentence ($p^-$). This technique is commonly employed in BERT's sibling models, such as SBERT \cite{reimers2019sentence} and PhraseBERT \cite{wang2021phrase}. However, unlike SBERT and PhraseBERT, which utilize the L2 norm to calculate the differences between anchor-positive and anchor-negative pairs, we employ cosine similarity to compute these similarity scores. Our preliminary experiments indicate that using the L2 norm approach performs worse than utilizing cosine similarity. Specifically, this method involves embedding triplets—comprising an anchor, a positive, and a negative sentence—using BERT to produce embeddings ($p$, $p^+$, $p^-$). The training goal is then to minimize the triplet loss:
\[
    \mathcal{L}_{triplet} = \max(0, \epsilon - cosine(p, p^+) + cosine(p, p^-))
\]
where $\epsilon$ is a margin and is set to 1 in our experiments.

The triplet loss does not significantly enhance the AND case due to the data generation process. This is because the positive sentence is usually a modification of the anchor sentence, such as a paraphrase or a positional change of two constraints. Meanwhile, the negative sentence contains only one constraint. Consequently, the positive sentence's embedding tends to be more similar to the anchor sentence's embedding than the negative one's. However, it does aid with the NOT case as the negative sentence often has more word overlap, paraphrases, or synonyms of the anchor sentence's entities that users wish to exclude. This similarity is more so than the positive sentence. The triplet loss pushes the negative embedding further from the anchor's, and brings the positive sentence's embedding closer to the anchor's embedding, even if there is not much word overlap. 

Pushing the contextualized embeddings of the positive-anchor pair closer together can be counterintuitive when considering the concept of similarity, especially when dealing with negation. For example, how should the query ``Trees not found in the Amazon'' be contextualized to be more similar to a document on ``Trees found in the Amazon'' compared to a document on ``Plants found in Europe''? This raises important questions about the nature of \textit{contextualized similarity}.

Negation plays a critical role in altering the context and meaning of queries and documents. In the query ``Trees not found in the Amazon,'' the negation fundamentally changes what the retrieval system should consider as similar. The document ``Trees found in the Amazon'' directly contradicts the query due to the presence of the negation, yet, there remains a conceptual connection—they both are about ``trees'' and ``the Amazon.'' Thus, while the negation creates a refutation, a degree of similarity persists because of the shared context around the topic of trees and the Amazon region.

However, this similarity is nuanced. It is vital that similarities are not ignored, but rather appropriately balanced so that the embeddings reflect both the shared context and the critical difference introduced by the negation. Negation should indeed push the embeddings apart to some extent, indicating differing meanings, but not to an extreme where they are treated as entirely dissimilar (as what the triplet loss does). This balance ensures that the retrieval system respects the contextual cues while retaining an understanding of the broader topic. Our experimental results will demonstrate that \textit{fine-tuning BERT using triplet loss degrades its performance on retrieval tasks involving logic-structured queries}.

\subsubsection{Inversed-Contrastive learning} Contrastive learning is a technique often used in dense retrieval approaches~\cite{izacard2021unsupervised,karpukhin2020dense}. The goal of contrastive learning is to retrieve the embedding of the document of interest among all negatives. Let $\mathcal{T}= \{\langle p_i, p^+_i, p^-_{i,1},...,p^-_{i,n}\rangle\}^m_{i=1}$ be a training data consisting of $m$ samples. Each sample contains an embedding of anchor sentence $p_i$, an embedding of sentence of interest $p^+_i$ and $n$ irrelevant (negative) sentences $p^-_{i,j}$. The contrastive learning's goal is to minimize the loss:
\begin{equation}
    \mathcal{L}(p_i, p^+_i, p^-_{i,1},...,p^-_{i,n}) = -\log \frac{e^{sim(p_i, p^+_i)}}{e^{sim(p_i, p^+_i)} + \sum_{j=1}^n{e^{sim(p_i,p^-_{i,j})}}}
\end{equation}
where $sim(\cdot)$ denotes a similarity metric of corresponding vectors, we use dot product in our paper. 

As stated earlier, our goal is to decrease unnecessary entity retrieval in instances involving NOTs (as well as ANDs). Hence, our focus is on the negative sentence. Rather than maximizing the similarity score for anchor-positive pairs, our objective is to minimize it for anchor-negative pairs. To increase the number of training examples, we also utilize \textit{in-batch negatives}. These are the positive and negative sentences in the same mini-batch that are not connected to the anchor question of interest. Consequently, the inversed-contrastive loss becomes:
\begin{equation}
    \mathcal{L}_{inv-contrastive} = -\log \frac{e^{-sim(p_i, p^-_i)}}{e^{-sim(p_i, p^-_i)} + \sum_{k=1}^l{e^{-sim(p_i,p^+_{i,k})}} + \sum_{j=1}^n{e^{-sim(p_i,p^-_{i,j})}}}
\end{equation}
where $l$ is the number of positive sentences of the anchor and $n$ is the number of in-batch negatives. In all our experiments, we use 2 positive sentences for every anchor sentence in the contrastive loss.

The final training goal of SetBERT is to minimize the inversed-contrastive loss $\mathcal{L}_{inv-contrastive}$.
\subsection{Implemetation details} The generated dataset is divided using an 80-20 ratio for training and evaluation. We fine-tune SetBERT on two NVIDIA Quadro RTX 8000 GPUs for 10 epochs, selecting the model with the smallest loss value on the evaluation set. Our batch size is 64 for Base configuration and 32 for Large configuration, and we use Adam for optimization with a learning rate of $1e-5$. 

\section{Experimental Setup}
We evaluate the efficiency of SetBERT on the task of document retrieval, specifically focusing on queries that contain boolean operators such as `and', `or', and `not'. Our evaluation is concentrated on the first-stage retrieval process, which involves identifying a broad set of relevant documents from a large corpus. The primary goal is to assess how well SetBERT handles complex query structures and logical conditions at this stage. 

We employ a benchmark dataset that includes diverse queries to test the model's ability to retrieve documents that match the specified boolean criteria. The re-ranking step, which involves fine-tuning the order of the retrieved documents based on their relevance to the query, is not included in this study. This exclusion allows us to isolate and measure the performance of SetBERT in the initial retrieval phase. Future work may explore how SetBERT performs during the re-ranking phase, as well as its integration with other retrieval and ranking models. By examining SetBERT's performance across different stages of the retrieval pipeline, we aim to provide a comprehensive understanding of its capabilities and potential areas for improvement.
\subsection{QUEST dataset}
We choose QUEST~\cite{malaviya2023quest} dataset because it offers a comprehensive benchmark for retrieval tasks involving queries with implicit set operations. QUEST comprises 325,505 entities, each with a document extracted from Wikipedia. The task involves retrieving one or more entities based on a given query. It includes 1307 queries for training, 323 for validation, and 1727 for testing. For more details about the dataset, please refer to the original paper.
\subsection{Metrics and Baselines}
\paragraph{Dual-Encoder Retriever}We adopt a dual-encoder framework as introduced in \cite{karpukhin2020dense} for our retrieval approach. This framework employs two distinct encoders: one dedicated to processing queries and another to processing documents, each trained to generate embeddings that capture semantic information from their respective textual inputs. Document relevance scores for a given query are computed using the dot product of the embeddings. The retrieval process selects the top-K documents with the highest computed scores. Comprehensive configurations and additional hyperparameter details can be found in the appendix \ref{apd: DE}.

Since we are focusing on the first-stage of retrieval, we use Recall@K - the proportion of relevant documents successfully retrieved out of the total number of relevant documents available, and MRecall@K~\cite{min2021joint} as the performance metrics. MRecall@K evaluates retrieval success by requiring all answers to be found when few, or at least $K$ answers to be found when there are many. We compare the dual-encoder (DE) initiated with SetBERT (denoted as SetBERT-DE) with sparse BM25 retriever \cite{robertson2009probabilistic} and DE initiated with BERT (denoted as BERT-DE). The DE framework setup follows DPR~\cite{karpukhin2020dense}.

\section{Evaluation results}
\subsection{Retrievers comparison}
In the comparative analysis of retrieval task performance metrics, it is evident that SetBERT models, both in their Base and Large configurations,  outperform their BERT counterparts by considerable margins. The performance metrics, Average Recall@k and Average MRecall@k, across various 
k values (20, 50, 100, 1000) (shown in Table \ref{tab: recall}), highlight the superiority of SetBERT.
\begin{table}[]
\caption{Performance in retrieval tasks: SetBERT-Base DE (blue) outperforms BERT-Base DE and achieves comparable results to BERT-Large DE, despite being three times smaller.}
\label{tab: recall}
\begin{tabular}{llllllllll}
\hline
                 & \multicolumn{4}{c}{Ave. Recall@k}               &  & \multicolumn{4}{c}{Ave. MRecall@k}              \\ \cline{2-5} \cline{7-10} 
Retriever        & 20    & 50    & 100            & 1000           &  & 20    & 50    & 100            & 1000           \\ \hline
BM25             & 0.104 & 0.153 & 0.197          & 0.395          &  & 0.020 & 0.030 & 0.037          & 0.087          \\
BERT-Base DE     & 0.082 & 0.134 & 0.182          & 0.453          &  & 0.006 & 0.011 & 0.017          & 0.086          \\
SetBERT-Base DE  & \textcolor{blue}{0.139} & \textcolor{blue}{0.220} & \textcolor{blue}{0.296}          & \textcolor{blue}{0.612}          &  & 0.016 & 0.032 & 0.050          & 0.207          \\ \hline
BERT-Large DE    & \textbf{0.146} & \textbf{0.227} & 0.300          & 0.627          &  & \textbf{0.021} & \textbf{0.041} & \textbf{0.062}          & 0.235          \\
SetBERT-Large DE & 0.145 & 0.223 & \textbf{0.314} & \textbf{0.653} &  & 0.016 & 0.034 & 0.058 & \textbf{0.237} \\ \hline
\end{tabular}
\end{table}

\textbf{BERT-Base DE underperforms BM25} with lower Average Recall and significantly worse Average MRecall at lower k values (e.g., 20 and 100). Even though BERT-Base DE shows some improvement in large-scale retrieval (higher k values), this does not outweigh its deficiencies in early retrieval precision.

\textbf{BERT-Base DE vs. SetBERT-Base DE} SetBERT-Base DE demonstrates a notable improvement over BERT-Base DE, with an Average Recall@1000 of 0.612 compared to BERT-Base's 0.453, reflecting a performance increase of approximately 35\%. Similarly, at k=100 SetBERT-Base achieves an Average Recall of 0.296 versus 0.182 for BERT-Base, indicating an improvement of about 63\%. In terms of Average MRecall, SetBERT-Base DE also outperforms BERT-Base DE significantly, with values of 0.207 and 0.086 at k=1000 respectively, showing a margin of over 140\%.

\textbf{BERT-Large DE vs SetBERT-Large DE} The performance gap is evident when comparing SetBERT-Large DE to BERT-Large DE. Although SetBERT-Large DE lags slightly at smaller values of k, it surpasses BERT-Large DE as k increases, particularly beyond k = 100. Specifically, SetBERT-Large DE achieves an Average Recall@1000 of 0.653 compared to 0.627 for BERT-Large DE, representing a relative improvement of approximately 4\%. While the performance margin is smaller compared to the Base versions, this improvement is consistently reflected across other metrics. At k=100, SetBERT-Large DE demonstrates an Average Recall of 0.314, outperforming BERT-Large DE's 0.300 by about 4.7\%. Additionally, in terms of Average MRecall, SetBERT-Large DE records 0.237 at k=1000 compared to 0.235 for BERT-Large DE. Although this difference is marginal, it consistently indicates superior performance by SetBERT-Large DE.

\textbf{SetBERT-Base and BERT-Large} demonstrate similar levels of efficacy, despite the significant difference in their parameter sizes. SetBERT-Base, with only 110 million parameters, showcases a performance that rivals BERT-Large, which comprises 336 million parameters, highlighting SetBERT's efficiency and effectiveness. For example, at k=1000, SetBERT-Base reaches an Average Recall of 0.612, compared to BERT-Large's 0.627. The difference here is around 2.4\%, showcasing SetBERT-Base’s competitive performance despite having a third of the parameters. The ability of SetBERT-Base to perform at nearly the same level as BERT-Large, with significantly fewer parameters, underscores its computational efficiency and makes it a more practical choice for applications where resources are limited.

To summarize, SetBERT models (both Base and Large) consistently outperform their BERT counterparts, indicating better retrieval performance. The improvements are more pronounced in the Large versions but still significant in the Base versions.
\subsection{Performance across query structures}
We evaluated SetBERT-DE, BERT-DE, and BM25 on seven distinct query structures, capturing the nuances of their retrieval capabilities. The query structures included simple single-condition queries, as well as more complex combinations of AND, OR, and NOT conditions. Table~\ref{tab: templates} (left) illustrates the Recall@1000 for each query structure across the three models.
\begin{table}[]
\caption{Recall@1000 across templates. Except for BM25, others use DE initiated by corresponding BERTs. -L is short for Large, -TL is short for training with Triple Loss. Left: BM25 and BERT-base configurations, right: BERT-Large configurations.}
\label{tab: templates}
\begin{tabular}{lccc|ccc}
\cline{1-5} \cline{7-7}
Template      & BM25  & \multicolumn{1}{l}{BERT} & \multicolumn{1}{l|}{SetBERT} & \multicolumn{1}{l}{BERT-L} & SetBERT-L & \multicolumn{1}{l}{SetBERT-L-TL} \\ \hline
A             & 0.395 & 0.470                    & \textbf{0.639}               & 0.625                          & \textbf{0.662}            & 0.562                             \\
A or B        & 0.535 & 0.497                    & \textbf{0.673}               & 0.669                          & \textbf{0.711}            & 0.623                             \\
A and B       & 0.313 & 0.424                    & \textbf{0.593}                        & 0.613                 & \textbf{0.636}            & 0.493                             \\
A not B       & 0.330 & 0.455                    & \textbf{0.626}               & 0.613                          & \textbf{0.644}            & 0.561                             \\
A or B or C   & 0.501 & 0.523                    & \textbf{0.655}                        & 0.669                 & \textbf{0.696}            & 0.600                             \\
A and B and C & 0.301 & 0.434                    & \textbf{0.593}                        & 0.640                 & \textbf{0.667}            & 0.547                             \\
A and B not C & 0.201 & 0.361                    & \textbf{0.495}                        & \textbf{0.550}                 & 0.548            & 0.482                             \\ \hline
\end{tabular}
\end{table}

SetBERT consistently outperforms BERT across all templates, with the most significant performance improvement observed in the ``A or B'' template and the smallest improvement in the ``A and B not C'' template. SetBERT achieves its highest Recall@1000 in the ``A or B'' query structure, with a score of 0.673. This performance is significantly higher than BERT's 0.497 and BM25's 0.535, indicating that SetBERT is particularly adept at handling inclusive OR conditions, where the query returns documents that match either of the specified terms. On the other hand, the ``A and B not C'' template is where SetBERT shows its lowest Recall@1000, with a score of 0.495. Despite being its least effective structure, SetBERT still outperforms BERT (around 37.7\%) and BM25 (around 146\%), demonstrating its capability to handle complex queries involving both inclusion and exclusion criteria. This comparative analysis underscores SetBERT's enhanced capabilities over BERT, making it a more effective model for a wide range of query structures in retrieval tasks.

\subsection{Contrastive loss versus Triplet loss}
In this section, we highlight the inefficiency of triplet loss when fine-tuning BERT for retrieval task with logic-structured queries. Table~\ref{tab: templates} (right) illustrates how training with triplet loss hinders performance of SetBERT. 

SetBERT-L-TL generally underperforms compared to both BERT-Large and SetBERT-Large, especilly for the intersection templates. In the ``A and B'' template, SetBERT-L-TL records 0.493, trailing behind BERT-L's 0.613 and SetBERT-L's 0.636 by 19.6\% and 22.5\%, respectively. This trend continues across other templates, with SetBERT-L-TL showing notable performance drops, particularly in complex conjunctive queries like ``A and B and C,'' where it scores 0.547, compared to BERT-L's 0.640 and SetBERT-L's 0.667. Additionally, SetBERT-L-TL underperforms the SetBERT Base configuration, which has consistently higher scores, such as 0.639 in the ``A'' template and 0.673 in the ``A or B'' template. These results suggest that the triple loss training approach may hinder SetBERT-L-TL's effectiveness, particularly in handling complex query structures, making it less efficient than both its large and base counterparts.

\section{Conclusion}
In this paper, we introduced SetBERT, a fine-tuned BERT-based model specifically designed to enhance query embeddings for set operations and boolean logics. Our work addresses the significant challenges faced by both traditional and neural retrieval methods in accurately interpreting and processing logic-structured queries. 

Our experimental results demonstrate that SetBERT significantly improves retrieval performance for these types of queries, making it a powerful tool for complex information needs. Moreover, we showed that fine-tuning with triplet loss, commonly used in models like SBERT and PhraseBERT, is not effective for this task. Instead, we proposed using an inversed-contrastive loss focusing on identifying the negative sentence to achieve better performance.

Notably, our SetBERT-base model is built by extending the smaller BERT model, it achieves results comparable to BERT-large despite being just one-third the size, highlighting the efficiency and effectiveness of our approach. These findings underscore the potential of SetBERT to advance the state-of-the-art in information retrieval, particularly for queries involving Boolean logic and set operations. Furthermore, this efficiency makes SetBERT-Base a valuable model for scenarios requiring high performance and resource efficiency. For future work, we aim to explore the applicability of SetBERT in domains beyond text retrieval, such as image and video retrieval, where Boolean logic queries could also provide significant benefits. 

\begin{acks}
This material is based upon work supported by the National Science Foundation under Award \# OIA-1946391, ``Data Analytics that are Robust and Trusted (DART)''.
\end{acks}

\bibliographystyle{ACM-Reference-Format}
\bibliography{bibfile}

\appendix
\section{Data Generation by Prompting}\label{apd: PROMPT}
\subsection{OpenAI API configurations}
Model: gpt-3.5-turbo; temperature: uniformly random between 0.5 and 1; top\_p: uniformly random between 0.8 and 1; max\_tokens=512, num\_samples\_per\_call=3, frequency\_penalty=0, presence\_penalty=0. Please refer to \url{https://openai.com/api} for more details.
\subsection{NOT}
As a helpful assistant, your role is to generate a dataset to fine-tune BERT using boolean NOT logic. Keep in mind that ``A not B'' means ``A and something different from B,'' such as ``C.'' Your task involves generating sentences following these guidelines:
\begin{enumerate}
    \item Gold Sentence: This sentence should have the form ``A that are not B'' or ``A but not B.''
    \item Positive Sentence: This sentence should be ``A that are C'' (where C is something different from B).
    \item Negative Sentence: This sentence should be ``A that are B.''
\end{enumerate}
\noindent For example:

Gold: ``Movies about Vietnam but not about war''

Positive: ``Films that are set in Vietnam or relate to Vietnamese culture'', ``films that are set in Vietnam or relate to Vietnamese history, society''

Negative: ``Movies set in Vietnam depict how cruel the war'',
``This movie is about Hanoi aerial war''.

\subsection{AND}
You are a helpful assistant who can help us generate a dataset to fine-tune BERT using boolean AND logic. Keep in mind that ``A and B'' can be ``B and A'' or their paraphrases, but cannot be just ``B'' or just ``A.'' Your task involves generating sentences in the following format:
\begin{enumerate}
    \item Gold: This sentence should have the form ``A and B.''
    \item Positive: This sentence should be either ``B and A'' or a paraphrase of ``A and B'' / ``B and A.''
    \item Negative: This sentence should be: ``A and something different than B'', ``B and something different than A'', just ``A'', just ``B''.
\end{enumerate}
\noindent{For example}:

Gold: ``British historical dramas that are  also 1960s historical films''

Positive: ``1960s historical films that are also  British historical drama films''

Negative: ``British historical drama films that are also 1980s historical films'', ``1960s historical films that are also American historical drama films''

\subsection{OR}
You are a helpful assistant who can help us generate a dataset to fine-tune BERT using boolean OR logic. In this context, ``A or B'' can mean either ``B'' or ``A''. Your task is to generate sentences following these guidelines:
\begin{enumerate}
    \item Gold Sentence: This sentence should have the form ``A or B''. 
    \item Positive Sentence: This sentence should be either ``B'' or ``A''.
    \item  Negative Sentence: This sentence should be something entirely different from both A and B.
\end{enumerate}

\noindent{For example:}

Gold: ``Non-fiction books about the Great Recession or Books by Matt Taibbi''.

Positive: ``Non-fiction books about the Great Recession'', ``Books by Matt Taibbi''.

Negative: ``Fiction books about the Great Recession'', ``Books by Mark Taiwan''.
\section{Dual-Encoder Hyperparameters} \label{apd: DE}
\begin{enumerate}
    \item Max query length: 64
    \item Max document length: 256
    \item Negative document per sample: 5
    \item Positive document: 1 (we use the first positive document)
    \item Batch size: 16
    \item Optimizer: Adam, learning rate: $1e-5$, linear scheduler
    \item Training steps: 1600
    \item Eval steps: 200, keep best
\end{enumerate}

\end{document}